\pdfoutput=1

\documentclass[11pt]{article}

\usepackage{acl}

\usepackage{times}
\usepackage{latexsym}
\usepackage{amsfonts,amssymb}
\usepackage{multirow}
\usepackage{multicol}
\usepackage{dirtytalk}
\usepackage{subcaption}
\usepackage{amsmath}
\usepackage{arydshln}
\usepackage{tikz}
\usetikzlibrary{shapes, positioning}
\usetikzlibrary{shapes}
\usepackage{pgfplots}
\usepackage{verbatim}
\usepackage{graphicx}
\usepackage{adjustbox}

\usepackage[T1]{fontenc}

\usepackage[utf8]{inputenc}

\usepackage{microtype}

\title{KERL: \textit{K}nowledge-\textit{E}nhanced Personalized Recipe \textit{R}ecommendation using \textit{L}arge Language Models \thanks{This paper has been accepted at the ACL 2025.}
}

 \author{Fnu Mohbat \and Mohammed J. Zaki \\
 Rensselaer Polytechnic Institute \\
  \texttt{mohbaf@rpi.edu}, 
  \texttt{zaki@cs.rpi.edu}
 }
 
\pgfplotsset{compat=1.18}
\begin{document}
\maketitle
\begin{abstract}

Recent advances in large language models (LLMs) and the abundance of food data have resulted in studies to improve food understanding using LLMs. Despite several recommendation systems utilizing LLMs and Knowledge Graphs (KGs), there has been limited research on integrating food related KGs with LLMs. We introduce KERL, a unified system that leverages food KGs and LLMs to provide personalized food recommendations and generates recipes with associated micro-nutritional information. Given a natural language question, KERL extracts entities, retrieves subgraphs from the KG, which are then fed into the LLM as context to select the recipes that satisfy the constraints. Next, our system generates the cooking steps and nutritional information for each recipe. To evaluate our approach, we also develop a benchmark dataset by curating recipe related questions, combined with constraints and personal preferences. Through extensive experiments, we show that our proposed KG-augmented LLM significantly outperforms existing approaches, offering a complete and coherent solution for food recommendation, recipe generation, and nutritional analysis. Our code and benchmark datasets are publicly available at \url{https://github.com/mohbattharani/KERL}.
 
\end{abstract}

\section{Introduction}

The importance of food for well-being has created the need to employ machine learning to promote healthy lifestyles through food understanding. Several recipe-sharing websites have created rich resources of food data, attracting researchers to devise food computing for classification, retrieval, recipe generation, and recommendation. Food recommendation is a complex and multifaceted taks given its  direct impact on human health. An effective food recommendation system should consider personal preferences, dietary constraints, and health guidelines. In recent years, several ontologies and knowledge graph methods have helped to better organize food data \cite{dooley2018foodon, haussmann2019foodkg, razzaq2023evorecipes}. Subsequently, several food recommendation methods have leveraged the KGs for personalized food recommendation \cite{chen2021personalized, shirai2021healthy, ling2022following, li2023health, kobayashi2024functional}. Several studies have also utilized LLMs for recipe generation \cite{h2020recipegpt, yin2023foodlmm, LlaVAChef} and nutrition estimation \cite{yin2023foodlmm, tanabecaloriellava, tanabe2025calorievol}.  However, there is a lack of unified food understanding systems that not only recommend personalized recipes but also generate cooking steps and micro-nutrition information for the recommended dishes. 
 



\begin{figure*}[ht]
    \centering
    \includegraphics[width=\textwidth,height=2.25in]{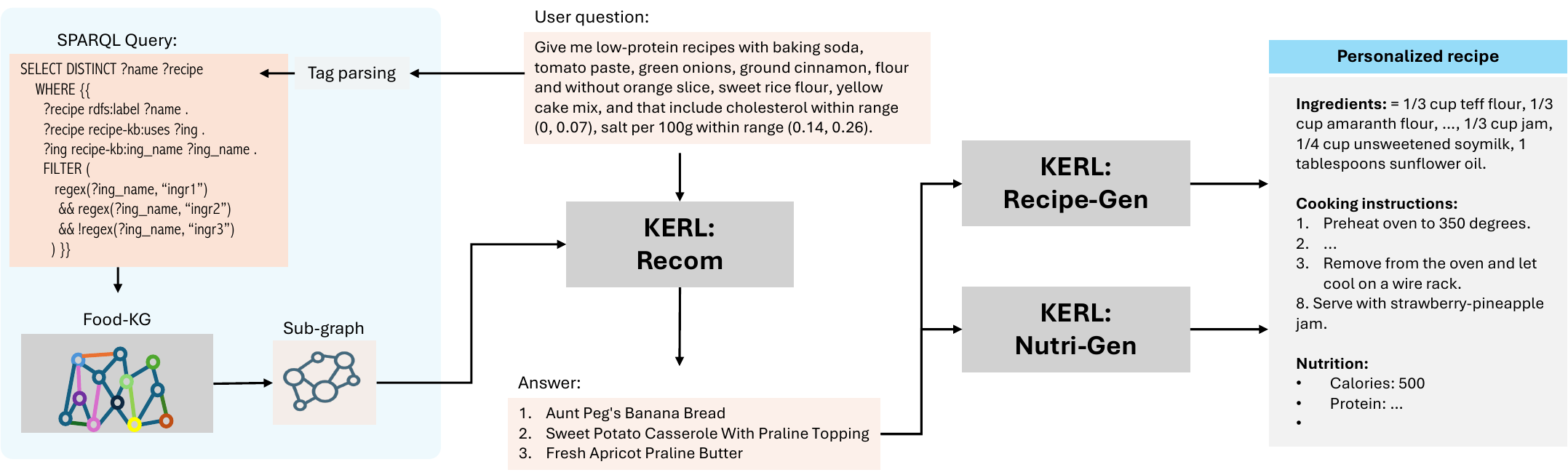}
    \caption{\small KERL Overview: Given a natural language question (with constraints), the system parses entities and generates a SPARQL query to retrieve a subgraph from the KG. The question and this subgraph as context, are given as input to the recommendation model (\textit{KERL-Recom}), which generates a list of recipe names that satisfy the constraints. The \textit{KERL-Recipe} and \textit{KERL-Nutri} models then generate cooking steps and micro-nutrients.}
    \label{fig:model1}
\end{figure*}


Despite the success of LLMs in multiple domains \cite{wu2023bloomberggpt, moor2023med, chhikara2024fire, LlaVAChef}, they are prone to hallucination and outdated information \cite{xu2024hallucination}. Retrieval-augmented generation (RAG) addresses the issue by utilizing documents or KGs as external knowledge \cite{mathur2024doc, he2024g, rangel2024sparql}. Question-answering over KGs (KGQA) retrieves the relevant subgraphs from KG, uses reasoning to extract entities as answers \cite{wang2021literatureqa}, or uses semantic parsing or LLMs in a zero or few shot setting to transform questions into SQL or SPARQL queries to get answers from the KG \cite{banerjee2023dblp, taffa2023leveraging, avila2024experiments}.  Despite various attempts to integrate external knowledge with LLMs in several domains, there is a lack of work on food recommendation that combines KGs and LLMs while considering both health constraints and user preferences.



We propose a personalized and unified food recommendation system called KERL that uses the FoodKG \cite{haussmann2019foodkg} as the knowledge source. The system illustrated in Fig.~\ref{fig:model1} comprises three modules: a recommendation module (\textit{KERL-Recom}), a recipe generation module (\textit{KERL-Recipe}), and a nutrition generation module (\textit{KERL-Nutri}), which are trained using a low-rank adaption (LoRA) approach~\cite{hu2022lora}.  The \textit{KERL-Recom} module takes a user query, extracts entities, constructs a SPARQL query to retrieve relevant subgraphs from the KG, and inputs these subgraphs along with the query into the LLM, which returns dish names that satisfy the constraints. The \textit{KERL-Recipe} modules generates recipes from the suggested titles, while the \textit{KERL-Nutri} produces detailed micro-nutritional information for recommended dishes. Overall, we make the following contributions.
\begin{itemize}
    
    \item We propose KERL, a unified food recommendation system based on a multi-LoRA approach, with a dedicated adapter for each task, while utilizing the same base model, allowing efficient training and inference.

    \item Our work generates comprehensive nutritional information unlike previous approaches that focus mainly on one aspect, such as calorie count.

    \item  We curated two open benchmark datasets using template questions, nutrient constraints, and personal preferences. 
    
    \item  Through extensive experiments, we show that each module of KERL outperforms the baseline LLMs, showcasing the power of integrating KGs with LLMs.
\end{itemize}


\section{Related Work}

\paragraph{Food Recommendation}
The initial food recommendation systems formulated recommendation as a retrieval task by mapping the recipe components such as title, ingredients, and images into a common embedding space \cite{salvador2017learning,chen2018deep, wahed2024fine, li2022food}. Later, the focus shifted towards the use of food knowledge graphs. For example,  \cite{li2022food, gao2022food} used graph neural network to learn the user-recipe interactions in KGs, and \citet{chen2021personalized} proposed knowledge base question answering through information retrieval by  mapping questions and possible answers in a common embedding space. However, recent methods employ LLMs for food recommendations. For example, \cite{kirk2023comparison} investigated ChatGPT for nutrition questions, and \cite{geng2022recommendation, rostami2024food} use LLMs as a language processing engine in the food recommendation system. Despite considerable efforts to leverage KGs and LLMs  for developing food recommendation systems, there remains limited research on integrating food KGs to augment LLMs for more personalized food recommendation. Specifically, individual preferences, health considerations, and nutritional constraints within a unified framework have not been extensively explored.

\paragraph{Question Answering Over KGs}
Question answering over knowledge graphs refers to retrieving knowledge from a KG to answer queries. Initial studies parsed entities from a natural language question and generated SPARQL queries from templates to retrieve the answers \cite{shirai2021healthy, haussmann2019foodkg, rangel2024sparql}. Later, researchers used embeddings from LSTM or graph neural networks and framed the problem as a retrieval task~\cite{chen2021personalized, gao2022food, he2024g}. Recent methods explore the integration of KGs to improve LLMs for reasoning \cite{luo2023reasoning,sun2023think}, chatbots for customer service \cite{xu2024retrieval}, product recommendations \cite{eppalapally2024kapqa}, and food-related tasks \cite{qi2023foodgpt, hou2024enhancing, ma2024large, zhang2024mopi}. For instance, FoodGPT \cite{qi2023foodgpt} aims to enhance recipe generation, while \cite{zhang2024mopi} focuses on recommending foods based on their health effects. Nevertheless, the full potential of KG and LLM integration in food science remains underexplored \cite{min2022applications, ma2024large}, presenting a critical research opportunity.


\paragraph{Recipe Generation}
One line of research focuses on generating the recipe title from food images or ingredients from the title, and then generating the recipes \cite{reusch2021recipegm, chhikara2024fire}. Several efforts tried to generate recipes directly from inputs such as title, images, and ingredients \cite{farahani2023chef, yin2023foodlmm, LlaVAChef}. RecipeGPT~\cite{h2020recipegpt} fine-tunes GPT-2~\cite{radford2019language} while RecipeMC~\cite{tanejamonte} refines the generated recipes using Monte Carlo Tree Search. RecipeGM~\cite{reusch2021recipegm} and Chef Transformer~\cite{farahani2023chef} generate recipes from ingredients, while FIRE \cite{chhikara2024fire} predicts those ingredients from a given image or title. However, more recent methods explore end-to-end fine-tuning of LLMs and multi-modal models (MMMs) for recipe generation. FoodLMM~\cite{yin2023foodlmm} fine-tunes LISA~\cite{lai2023lisa} for classification, ingredient detection, segmentation, and recipe generation, while LLaVA-Chef \cite{LlaVAChef} investigates better fine-tuning schemes to improve recipe generation.  One recent work \cite{liu2024retrieval} even tried retrieval augmented generation (RAG) for recipe generation. However, all of these methods focus solely on recipe generation.  

\paragraph{Nutrition Generation}
Due to the effectiveness of nutritional intake for personal health, researchers employed MMMs for calorie estimation \cite{yin2023foodlmm, tanabecaloriellava,   yao2024caloraify, tanabe2025calorievol}  from food images. Most of these methods aim to estimate the calories from one or more food images utilizing the Nutrition5k \cite{thames2021nutrition5k} dataset that contains only 5000 recipes with a total of 125K images. FoodLMM \cite{yin2023foodlmm} leverages LISA~\cite{lai2023lisa}, CalorieLLaVA \cite{tanabecaloriellava} fine-tunes LLaVA \cite{liu2024visual} and CaLoRAify \cite{yao2024caloraify} fine-tunes  Llama-2 based visual language model for calorie estimation. Most of the existing work is limited to calorie estimation only, disregarding the estimation of other vital micro-nutrients. This work considers the estimation of several micro-nutrients, including protein, fiber, fat, and cholesterol.  



\begin{figure}[!ht]
    \centering
    \includegraphics[width=0.95\linewidth]{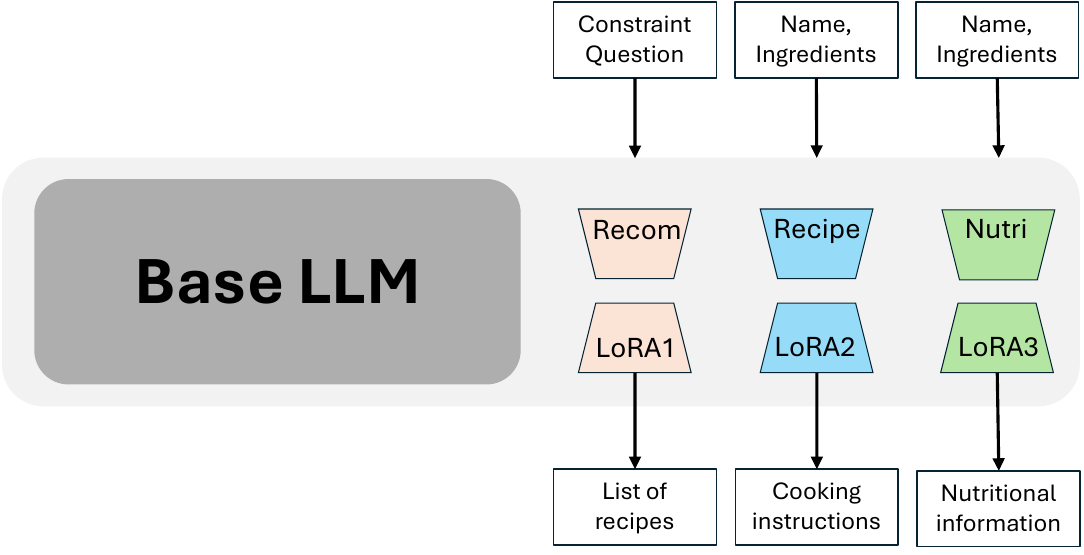}
    \caption{\small KERL Multi-LoRA Setup: With the same base model, a separate LoRA adapter is trained for each task. During inference, the desired adapter is activated while base model remains the same.}
    \label{fig:model_training}
\end{figure}

\section{KERL: Food Recommendation System}
We propose KERL, a personalize food recommendation system that unifies recommendation with cooking steps and nutrition details generation by leveraging multi-LoRA approach as illustrated in Fig.~\ref{fig:model1}. KERL uses the FoodKG \cite{haussmann2019foodkg} as external knowledge source and an LLM as a generative engine. FoodKG contains 1 million recipes from Recipe1M~\cite{salvador2017learning} with ingredients, nutritional information, and tags, organized into 67 million triplets. Let $t_j$ be a tag in FoodKG, and $R(t_j)$ the set of tagged recipes with tag $t_j$ where each recipe $X_{recipe}$ has a title or name $X_t$, ingredients $X_{ing}$, cooking steps $X_{inst}$ and nutrients $X_{nutri}$. Now, let $I(t_j)$ be the set of ingredients in all recipes in $R(t_j)$, then we define $I^+(t_j) \subset I(t_j)$ as the set of ingredients that the user likes to have and $I^-(t_j) \subset I(t_j)$ as the set of ingredients that the user wants to avoid. Let $X_{nutri,i}$ be the $i^{th}$ nutrient, then $\mu (R(t_j), X_{nutri,i})$ and $\sigma (R(t_j), X_{nutri,i})$ denote the mean and standard deviation of the nutrient, respectively. These statistical measures are later used to define nutrient-based preference filters. To incorporate these constraints into personalized recipe recommendations, we employ a modular approach in designing KERL's multi-LoRA architecture, as shown in Fig.~\ref{fig:model_training}, where each LoRA adapter is fine-tuned for a distinct sub-task. The \textit{KERL-Recom} adapter is fine-tuned to identify dishes from $R(t_j)$ that satisfy the constraints in the user query as recommended recipes. Subsequently, we also fine-tune the \textit{KERL-Recipe} and \textit{KERL-Nutri} adapters for generation of cooking instructions and micro-nutrients for the recommended recipes. Together, the three modules comprise a comprehensive and integrated food recommendation system.

\subsection{KERL-Recom}
Given a complex user query containing allowed and disallowed ingredients, and other nutrition-based constraints (see Table~\ref{tab:temp_questions} for some examples), the main task for the \textit{KERL-Recom} adapter is to recommend relevant recipes leveraging the FoodKG to return a high quality response.
The steps involve retrieval of query relevant subgraphs from the KG, fine-tuning the model on user queries, and model inference over the KG for generating the response as recommended recipes. 

\smallskip\noindent
\textbf{Subgraph Retrieval:}
Given a natural language question, the first step is to parse entities such as the tag $t_j$ (e.g., \textit{American}, \textit{Healthy}) and ingredients (e.g., \textit{sugar}, \textit{cheese}). These entities are then used to generate SPARQL queries based on predefined templates, allowing us to retrieve relevant subgraphs from FoodKG. Each subgraph contains the name of the dish, the list of ingredients, and nutritional information. The subgraphs are serialized into a text sequence and given to LLM as a context. An example of a KG subgraph, along with the relevant recipe text is shown in Fig.~\ref{fig:sample1} (See Appendix). 


\begin{table*}[!ht]
\centering
 \begin{adjustbox}{width=0.85\textwidth}
 \small
    \begin{tabular}{|p{0.49\textwidth}|p{0.49\textwidth}|}

\hline

\begin{description}
    \item [Base question:] Give me \textit{\{tag\}} recipes with \textit{\{ingredients\}} and without \textit{\{not\_have\_ingredients\}}
    \item [Template constraints:] have \textit{\{nutrition\}} no more than \textit{\{limit\}},   \textit{\{nutrition\}} within range  \textit{\{limit\}}
    \item [Personal preferences:] 
    \begin{description}  
    \item [tag:] low-protein
    \item [Likes:] baking soda, tomato paste, green onions, ground cinnamon, flour
    \item [Dislikes:] orange slice, rice flour, yellow cake mix
    \item [Nutrition constraints:] cholesterol no more than 0.07, salt per 100g (0.14, 0.26)
\end{description} 

    \item [Question:]Give me low-protein recipes with baking soda, tomato paste, green onions, ground cinnamon, flour and without orange slice, sweet rice flour, yellow cake mix, and have cholesterol no more than 0.07, salt per 100g within range (0.14, 0.26).
    \item [Answer:] Aunt Peg\'s Banana Bread,
        Sweet Potato Casserole With Praline Topping,
        Fresh Apricot Praline Butter. 
\end{description}
&

\begin{description}
    \item [Base question:] What are the \textit{\{tag\}} dishes that contain \textit{\{ingredients\}} but do not contain \textit{\{not\_have\_ingredients\}} 
    \item [Template constraints:] have \textit{\{nutrition\}} at least \textit{\{limit\}},  and \textit{\{nutrition\}} less than \textit{\{limit\}}
    \item [Personal preferences:] 
    
\begin{description}  
    \item [tag:] vegetarian
    \item [Likes:] margarine, frozen peas, shredded cheddar cheese, baking soda, vinegar
    \item [Dislikes:] cracked wheat, chili pepper, fresh pepper
    \item [Nutrition constraints:] fiber at least 4.24, saturated fat less than 6.49
\end{description}

    \item [Question:]  What are the top vegetarian recipes containing margarine, frozen peas, shredded cheddar cheese, baking soda, vinegar and excluding cracked wheat, chili pepper, fresh pepper, and meeting the fiber at least 4.24, saturated fat less than 6.49 condition?
    \item [Answer:]  B. B. King\'s German Chocolate Cake,
            Apple Bread,
            Mom\'s Raisin Rock Cookies 
\end{description}

        \\
\hline

\hline
\end{tabular}
\end{adjustbox}
    \caption{\small Examples of constraints, corresponding questions, and relevant recipe names as ground truth answers. Ingredient preferences specify whether certain ingredients should or should not be included in the recipes. Nutritional constraints are numerical conditions applied to nutrient values, defined by limits such as less than, greater than, or within a specified range.    
    }
    \label{tab:temp_questions}
\end{table*}
%

\smallskip\noindent
\textbf{Model Optimization: } \textit{KERL-Recom} model is trained to select recipes from the context that meet the constraints in the question. Given that $R(t_j)$ is the set of recipes with the relevant user tag $t_j$, let $R^+(t_j)$ denote the {\it positive} subset of recipes that meet all query constraints, and let $R^-(t_j)$ denote the rest of the recipes that make up the {\it negative} subset.  
During training, we select a subset of recipes of size at most $K$, such that we sample at most $K/2$ positive recipes from $R^+(t_j)$ and at most $K/2$ negative recipes from $R^-(t_j)$. This determines the context $C_j$ for the LLM training, along with the full query, with $K$ chosen so that the model can fit within the GPU memory.
This approach allows the model to learn to select from a wide distribution of contexts (positive or negative recipes) despite $|R^-(t_j)| \gg |R^+(t_j)|$ (e.g., see the dataset statistics in Table \ref{tab:dataset_stats}). 
The recipes sampled from $R^+(t_j)$ serve as the ground truth answer $Y$. The model is trained using low-rank adaptation \cite{hu2022lora}
(see details in Sec. \ref{apndx:training}) with standard cross-entropy loss: $ L_{CE} = CE (p (Y), p (\Tilde{Y}))$, where $p (Y)$ is probability of ground truth recipe tokens as one hot vector and $p (\Tilde{Y})$ is the predicted probability of the recipe tokens generated by the model.

\smallskip \noindent
\textbf{Inference over KG:}
During inference, the entire FoodKG could theoretically be the context for searching relevant recipes. However, in practice, we parse the tag $t_j$ from the query and retrieve $R(t_j)$ as context. The maximum number of tagged recipes could be potentially very large, and the resulting total number of tokens may exceed the LLM's sequence length, which may also lead to GPU memory overflow. Therefore, like in training, we iterate over $R(t_j)$ by providing the LLM with the query and a subset of $R(t_j)$ as context $C_j$, and combine the responses from multiple calls to the LLM to generate the final answer. This approach allows us to perform inference and evaluate the model on a variable number of recipe subgraphs.

\subsection{KERL-Recipe}
\textit{KERL-Recipe}, the recipe generation module enables the recommendation system to generate recipe steps. This module can leverage any recipe generation model, such as LLaVA-Chef~\cite{LlaVAChef} or FoodMMM~\cite{yin2023foodlmm}. While these models rely on older LLM backbones, recent advances such as LLaMA-3~\cite{meta2024introducing} and Phi-3~\cite{abdin2024phi} have significantly outperformed their predecessors. Therefore, we employed Phi-3-mini for recipe generation, specifically generating cooking steps from the dish names $X_t$, and ingredients $X_{ing}$, or both. Unlike ~\cite{LlaVAChef, yin2023foodlmm}, we use LoRA training, which reduces the number of training parameters and decreases the training time. Thus, KERL-Recipe, implemented as a LoRA adapter, integrates seamlessly into the KERL framework, while utilizing Phi-3 Mini as the base model.  


\subsection{KERL-Nutri}
\textit{KERL-Nutri}, the nutrition generation module, is also a LoRA adapter trained to generate micro-nutritional information from the recipe name $X_t$, the ingredients $X_{ing}$, and the cooking steps $X_{instr}$ or their combination. The module helps ensure that the recommended recipes follow the nutritional constraints in the user query. 


All three modules share the same backbone LLM (namely, Phi-3-mini~\cite{abdin2024phi}), with separate LoRA adapters fine-tuned for each task as illustrated in Fig.~\ref{fig:model_training}. Training details and hyperparameters are discussed in Sec. \ref{apndx:training}. This design allows multiple adapters to operate even on a single GPU, enabling practical and efficient inference.

\section{Benchmark Generation}

One of our contributions is the creation of  a large benchmark dataset of realistic constrained user queries for training and evaluation, given the lack of real user data. For \textit{KERL-Recom} and \textit{KERL-Nutri}, we curated base (template) questions using GPT-4 \cite{achiam2023gpt}, whereas for \textit{KERL-Recipe}, we borrowed template prompts from LLaVA-Chef \cite{LlaVAChef} (see Sec.~\ref{apendx:recipe_gen} in Appendix). Based on the task, each base question contains placeholders for inputs which are then replaced with their values.

\subsection{Generating Personal Preferences}

To personalize the recommendation of recipes, individualized information about the person’s likes, dislikes, and other personal choices are important. 
Our benchmark incorporates both ingredient preferences and nutritional constraints. 
We combine the base question and constraints to obtain the final constrained question. See Table~\ref{tab:temp_questions} for examples. Recipes that satisfy all constraints are considered ground truth answers or a positive set of recipes $R^+(t_j)$,  and the remaining $R^-(t_j) = R(t_j) - R^+(t_j)$ are considered as a negative set of recipes. This allows us to generate personalized food recommendations that take into account both taste preferences and dietary needs.


\smallskip\noindent
\textbf{Ingredient Preferences: }
Ingredient preferences consider what a recipe should or should not contain (e.g., \textit{Recipe should contain Spinach and Butter but must not have Nuts}). Given a set of tags, for each tag $t_j$  we create a set of ingredients $I(t_j)$ used by all recipes in $R(t_j)$. To model a person's likes and dislikes of ingredients, we randomly sample two mutually exclusive sets of ingredients $I^+(t_j)$ and $I^-(t_j)$ from $I(t_j)$ such that $I^+(t_j) \cap I^-(t_j)  = \emptyset $. One set $I^+(t_j)$  is treated as person's preferred ingredients, while the other set $I^-(t_j)$  is considered as disliked ingredients that one may wish to avoid in the recipes.

\smallskip\noindent
\textbf{Nutritional Constraints:}
We also generate nutrition-related user preferences by defining constraints on nutrients (e.g.,\textit{recipes with more than 2 grams of protein and less than 500 calories}). Each constraint is defined in the format: <nutrient> <limit> <value> (e.g., salt less than 0.5g). The limit can be one of three filters: \lq less than\rq, \lq greater than\rq, or \lq fall within a defined range\rq. The <value> represents the threshold for the limit, or a range.
To generate nutritional constraints, first we randomly select one of the three filters. Then, we define a threshold for the nutrient $x_i^{thresh}$ by sampling a random number in the range of $\mu (R(t_j), X_{nutri,i}) \pm 2 \sigma (R(t_j), X_{nutri,i})$, where $\mu$ and $\sigma$ are the mean and standard deviation. For the range filter, the upper and lower bounds are set as either $(0, x_i^{thresh})$ or $(x_i^{thresh}, max(X_{nutri,i}))$. 
Finally, all selected nutritional constraints are combined with the base question. This approach enhances the diversity of the questions, incorporating both conditional logic and negations, which are crucial for generating more complex and realistic queries.



\begin{table}[!ht]
    \centering
    \begin{adjustbox}{width=0.48\textwidth}
    \begin{tabular}{|l|c|c|c|c|}
    \hline
    \multirow{2}{*}{Measure}  & \multicolumn{2}{|c|}{\bf KGQA Benchmark} & \multicolumn{2}{|c|}{PFoodReq}\\
     & Train set & Test set &  Train set & Test set\\
      \hline
      Number of questions   & 62320 & 7790 & 4613 & 2305\\
       $R(t_j)$ (min)   &7& 7  & 2 & 2\\
       $R(t_j)$   (max)  &4445& 4445  & 2486 & 2485\\
       $R(t_j)$   (avg)   &3167& 3163 & 408.4 & 377.99\\
      \hdashline
       $R^+(t_j)$  (min)&  1 & 1 & 1& 1\\
       $R^+(t_j)$ (max)  &1776 & 954 &296 & 178 \\
       $R^+(t_j)$ (avg) & 10.67 & 9.77 & 2.94 & 2.84 \\
     
      \hline
    \end{tabular}
\end{adjustbox}
    \caption{\small KGQA Benchmark: Total number of questions, and the number of tagged recipes for overall context $R(t_j)$ and ground truth answer $R^+(t_j)$.}
    \label{tab:dataset_stats}
\end{table}


\subsection{KGQA Benchmark}
In the KGQA benchmark, ingredient preferences and nutrition constraints are combined with a user query to create a detailed question. Examples of the base question, constraints, nutritional limits, and the final question are given in Table \ref{tab:temp_questions}. To generate final queries, we randomly sample a base question from the templates, replace the placeholders with ingredient choices and nutritional constraints. The recipes that meet all the conditions in the final question are considered recommended recipes.

We used FoodKG as our knowledge base, containing over 1 million recipes labeled with 490 unique tags, where each recipe may have multiple tags. Questions were generated based on health-related tags, e.g., dairy-free, low-fat, high-fiber (full list of tags is in Appendix \ref{appendx-tags}). Our dataset consists of 77,900 question-answer pairs, split into $80\%$ for training, $10\%$ for validation, and $10\%$ for testing. Table \ref{tab:dataset_stats} shows that 
the number of recipes for a given tag $R(t_j)$, which is also the possible context size $|C_j|$, ranging from 7 to 4,445, while the recipes in the ground truth answer $R^+(t_j)$ vary from 1 to 954, highlighting the complexity and variety of the questions.
Note also, that our KGQA benchmark is over an order of magnitude larger than the pFoodReq dataset~\cite{chen2021personalized}, which has a total of only 6918 questions.

\begin{table}[!ht]
    \centering
    \small 
    \fbox{ 
    \begin{minipage}{0.93\linewidth} 
    \begin{itemize}
         \item For <name>, can you calculate the approximate nutritional values for a standard serving? 
        \item Estimated nutritional values for <name>. 
        \item Generate the nutritional values of the dish based on the ingredients: <ingredients>.
        \item A dish is cooked using <ingredients>, calculate the nutritional values of the dish.
        \item Generate the nutritional values of the dish based on its step-by-step instructions: <instructions>.
        \item Based on the cooking instructions provided, calculate the nutritional values of the dish. Instructions: <instructions>.
        \item For the following dish, estimate the nutritional values. Recipe: <name> <ingredients> <instructions>.
    \end{itemize}
    \end{minipage}
    }
        \caption{Example prompts utilized for training the \textit{KERL-Nutri} model, where placeholders were replaced with the corresponding information.}
    \label{tab:prompts_nutri_gen}
\end{table}

\subsection{Nutrition Generation Benchmark}
In the absence of a standardized benchmark for micro-nutrients, we sourced ground-truth micro-nutritional information from Recipe1M (and thus FoodKG) and \cite{li2023health}, resulting in about 500,000 recipe samples for which we were able to gather nutritional information. Using Recipe1M's predefined train-test splits, we use 19,000 recipes for our test set, with the remaining recipes used as the training set for our nutrition generation benchmark. Subsequently, to train LLMs, we curated about 40 template prompts using GPT-4, with examples of some of the prompts given in Table~\ref{tab:prompts_nutri_gen}. 
The placeholders <name>, <ingredients>, and <instructions> in the prompts are replaced with their corresponding actual information from the dataset samples. For example, in the prompt "Estimated nutrition for <name>" the placeholder <name> is replaced with the recipe title (name) $X_t$, which is then input to the LLM to generate the nutritional information. 
The prompts are intended to generate nutritional information from recipe attributes such as the title $X_t$, ingredients $X_{ing}$, and cooking instructions $X_{instr}$, or their combinations, which allows the model to learn nutritional information from different attributes of the recipes.




\subsection{Recipe Generation Benchmark}
\label{apendx:recipe_gen}

Recipe generation benchmark utilizes Recipe1M \cite{salvador2017learning}, which contains over 1 million recipes, split into train, test and validation sets. The training set consists of 720,639 recipes. For the test set, we use a filtered version of the Recipe1M test set from LLaVA-Chef \cite{LlaVAChef}, referred to as \texttt{test50k}, which contains 50,000 recipes. We base our approach on the template prompts used in LLaVA-Chef \cite{LlaVAChef}, which employed GPT-3.5 to generate these prompts. The prompts are designed to generate recipes from a given title ($X_t$), a list of ingredients ($X_{ing}$), or both. The example prompts are provided in Table \ref{tab:prompts_recipe_gen}.

\begin{table}[!ht]
    \centering
    \small
    \fbox{ 
    \begin{minipage}{0.95\linewidth} 
    \begin{itemize}
     \item Generate a comprehensive recipe for crafting <name>. 
     \item Detail the method for cooking a delightful <name>. 
     \item Construct a detailed cooking procedure for <name>.  
     \item Generate a recipe using <ingredients>.  
     \item Given <ingredients>, give the detailed recipe.
     \item Compose a recipe for making a dish using the ingredients: <ingredients>. 
     \item Generate a recipe for crafting <name> using <ingredients>.  
     \item Outline the process of making a delicious <name> using <ingredients> 
     \item Given <ingredients>, suggest me recipe of <name> 

    \end{itemize}
    \end{minipage}
    }
    \caption{\small Example prompts utilized training \textit{KERL-Recipe} model, where placeholders were replaced with the corresponding information.
    }
    \label{tab:prompts_recipe_gen}
\end{table}


\section{Experimental Results}


 
 For baseline comparison, we select several open source LLMs, as detailed Appendix \ref{apdx-subsec:llms}. We report the performance of \textit{KERL-Recom} on standard retrieval metrics such as precision, recall, and F1, and \textit{KERL-Recipe} on various text generation and summarization metrics including BLEU \cite{papineni2002bleu}, Rouge \cite{lin2004rouge}, METEOR ~\cite{elliott2013image} and CIDer \cite{vedantam2015cider}. For \textit{KERL-Nutri}, we parse micro-nutrients from the generated response and compute the mean average error (MAE) with ground truth. The metric definitions are provided in Appendix \ref{apdx-subsec:metrics}. Our code and benchmark datasets can be found at \url{https://github.com/mohbattharani/KERL}.

\subsection{Experimental Setup}

\label{apndx:training}
We leverage Phi-3-mini for its performance and compact size and fine-tuned one LoRA \cite{hu2022lora} adapter per task. For each task, we used the same LoRA configuration with $r=64$ and $\alpha = 16$ and $dropout = 0.5$, where $r$ is the dimensionality of the low rank and $\alpha$ is the scaling factor. We trained a separate LoRA adapter for each task, the overall model is shown in Figure \ref{fig:model_training}. During inference, the same model with multiple adapters allows us to deploy once and activate the task-specific adapter as needed. All experiments were performed on four NVIDIA RTX A6000 GPUs. Each LoRA adapter is trained for two epochs on task related dataset. The training hyper parameters were kept the same for all models, with a starting learning rate of $lr = 2 \times 10^{-5}$ and a cosine learning rate scheduler. During validation, hyperparameters were also fixed for all the models. Specifically,
we used \texttt{temperature = 0.2}, \texttt{num beams =1} and maximum new tokens to 1024.


\begin{table}[!ht]
\centering
 \begin{adjustbox}{width=0.48\textwidth}
\scriptsize
    
    \begin{tabular}{|l|c|c|c|c|c|}
    \hline
      Model      & mAP & P & R & F1  \\
      \hline
      internLM2 \cite{cai2024internlm2}       & 0.06 & 0.024 & 0.055 & 0.034   \\
      Mistral \cite{jiang2023mistral} &      0.214 & 0.536 & 0.558 & 0.547  \\
      Phi-2 \cite{mojan2023phi2} &       0.271 & 0.084 & 0.378 & 0.137  \\
      Llama-2 \cite{touvron2023llama2} &   0.557 & 0.825 & 0.627 & 0.713   \\
      Llama-3.1 \cite{meta2024introducing} &      0.146 & 0.28 & 0.406 & 0.332  \\
      Phi-3-mini-4K \cite{abdin2024phi} &     0.047 & 0.192 & 0.044 & 0.071   \\
      Phi-3-mini-128K \cite{abdin2024phi} &      0.275 & 0.778 & 0.278 & 0.41   \\
      \hline
      
       \textit{KERL-Recom} &  {\bf 0.96} & {\bf 0.978} & {\bf 0.969} & {\bf 0.973}  \\
      \hline
    \end{tabular}
  \end{adjustbox}

    \caption{\small KGQA Benchmark Test Set: \textit{KERL-Recom} versus  pre-trained LLMs. }
    \label{tab:test_8k_1}
\end{table}

\subsection{\textit{KERL-Recom} Evaluation}

\paragraph{\textit{Comparison with Open Source LLMs:}}
Table~\ref{tab:test_8k_1} presents the results of recent state-of-the-art LLMs for recipe recommendation. Despite claims of superiority by internLM2 \cite{cai2024internlm2} and Llama-3.1 \cite{meta2024introducing} on various benchmarks, both failed to understand the complex constraints in our KGQA benchmark questions. The capability of handing larger sequence length by Phi-3-mini-128K \cite{abdin2024phi} helps it perform better than Phi-3-mini-4K. Therefore, we selected Phi-3-mini-128K as the base LLM for KERL due to its compact size (3.8B parameters) and competitive performance. 
Our fine-tuned LoRA \textit{KERL-Recom} model significantly outperforms the other models, achieving a 56-point improvement over Phi-3-mini-128K and 26-point improvement over the larger Llama-2-7B \cite{touvron2023llama2} in F1 score.


\begin{figure*}[!ht]
    \centering
    \includegraphics[width=0.98\linewidth]{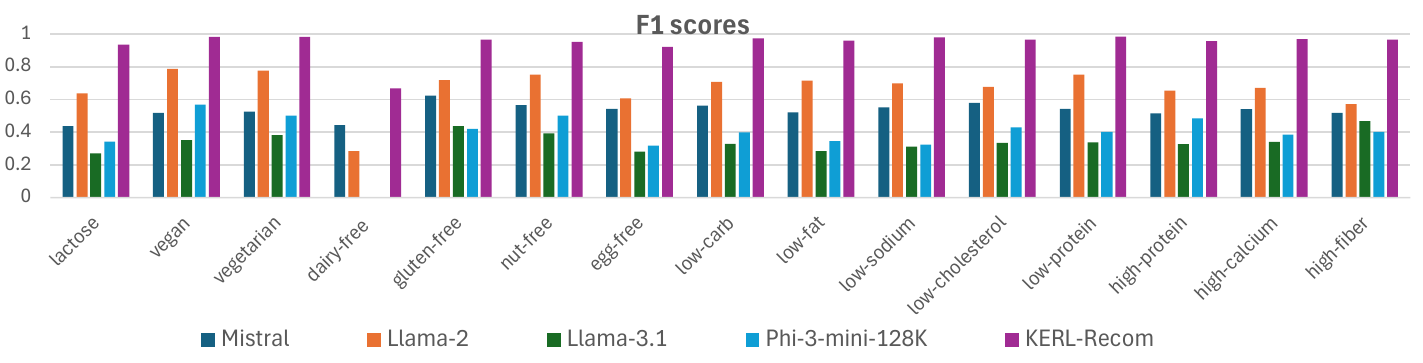}
    \caption{F1 scores of different models across various recipe types. Our model, KERL-Recom, consistently outperforms others with a significant margin in all categories.}
    \label{fig:per_tag_f1}
\end{figure*}


\begin{table}[!ht]
\centering
 \begin{adjustbox}{width=0.42\textwidth}
    \scriptsize
    \begin{tabular}{|l|c|c|c|c|}
    \hline
      Tag &  mAP & P & R & F1  \\
      \hline
      lactose   & 0.898 & 0.955 & 0.916 & 0.935 \\
      vegan       & 0.964  & 0.988 & 0.975 & 0.981\\
      vegetarian  & 0.966& 0.987  & 0.976 & 0.981 \\
      dairy-free   & 0.667 & 0.667   & 0.667 & 0.667\\
      gluten-free   & 0.922 &  0.992 & 0.779 & 0.873\\
      nut-free &   1.0 & 0.909 & 1.0 & 0.952 \\
      egg-free &   0.939 & 0.982  & 0.951 & 0.966\\
      low-carb &   0.952  & 0.983 & 0.965 & 0.974\\
      low-fat&   0.964 & 0.956 & 0.966 & 0.961\\
      low-protein&   0.981 & 0.988  & 0.981 & 0.984\\
      low-sodium&   0.982 & 0.978 & 0.982 & 0.98\\
      low-cholesterol  & 0.924& 0.98  & 0.951 & 0.965\\
      high-protein&   0.992 &  0.944 & 0.967 & 0.956 \\
      high-calcium&   0.937 & 0.981 & 0.953 & 0.967 \\
      high-fiber& 0.938 & 1.0  & 0.933 & 0.966 \\
      
      \hline
    \end{tabular}
    \end{adjustbox}
    \caption{\small \textit{KERL-Recom}: per tag results on KGQA test set}
    \label{tab:test_8k_per_tag}
\end{table}

\paragraph{\textit{Impact of Recipe Types:}}
To evaluate the generalization across various types of recipes, we compare F1 score for the baseline models and \textit{KERL-Recom} in Fig.~\ref{fig:per_tag_f1} (and Table \ref{tab:test_8k_2} in Appendix); our model consistently outperforms all others. For \textit{dairy-free} recipes, only Llama-2 and our model could recommend the correct recipes. Furthermore, the similarly high scores for \textit{KERL-Recom} for most recipe types, except for \textit{dairy-free} and \textit{gluten-free}, as shown in 
Table~\ref{tab:test_8k_per_tag} indicates that it generalizes well across different types of dishes.
Relatively lower accuracy and F1 scores for \textit{dairy-free} recipes is due to the dearth of training samples of this tag (see Table~\ref{tab:kgqa_details} in Appendix). 

\paragraph{\textit{Comparison on pFoodReq Benchmark:}}
The PFoodReq approach \cite{chen2021personalized} can also generate recommendations for constrained queries. However, it does it via an embedding-based approach that computes the similarity between the user query embedding and KG subgraph embeddings. 
Table \ref{tab:kbqa_results} shows how our KG-enhanced LLM approach performs on the pFoodReq benchmark dataset. We see that \textit{KERN-Recom} outperforms pFoodRec by 21.7 points on F1; it also outperforms Llama-2-7B by 56.5 points. Our method by utilizing the power of generative models generalizes better and outperforms the classical embedding based methods. 

\begin{table}[!ht]
\centering
 \begin{adjustbox}{width=0.38\textwidth}
\small
 \begin{tabular}{|l|c |c|c|c| }
    \hline
      Model     & mAP & P & R & F1   \\
      \hline
    P-MatchNN   & 0.455 & - & 0.451 & 0.412  \\
    pFoodReq    & 0.627 & - & 0.618 & 0.637     \\
    Llama-2 &   0.322 & 0.204 & 0.498 & 0.289 \\
    \hline
    \textit{KERL-Recom}   & {\bf 0.769} & {\bf 0.825} & {\bf 0.885} & {\bf 0.854}    \\
      \hline
    \end{tabular}
\end{adjustbox}
    \caption{\small PFoodReq Results: Our model compared to baseline shows better scores.}
    
    \label{tab:kbqa_results}
\end{table}


    


\begin{table*}[!htb]
    \centering
 \begin{adjustbox}{width=0.98\textwidth}
    \begin{tabular}{|l|l|c|c|c|c|c|c|c|c|c|c|c|}
\hline

Model & Inputs & BLEU-1 & BLEU-2 & BLEU-3 & BLEU-4 & SacreBLEU & METEOR & ROUGE-1 & ROUGE-2 & ROUGE-L  & CIDEr & Perplexity $\downarrow$ \\
\hline
LLaMA~\cite{touvron2023llama2} & $X_t$ + $X_{ing}$ & 0.252 & 0.129 &0.072 & 0.043 & 0.053 & 0.156 & 0.293 & 0.077 & 0.156 & 0.031 & 2.86   \\

LLaVA~\cite{liu2024visual} & $X_{i}$ + $X_{t}$ + $X_{ing}$ &  0.290 & 0.155 & 0.087 & 0.051 & 0.06 & 0.20 & 0.366 & 0.105 &  0.184 & 0.041 & 12.39 \\

\hdashline

LLaVA-Chef  & $X_t$ & 0.283 & 0.149 & 0.081 & 0.047 & 0.116 & 0.142 & 0.37 & 0.108 & 0.193 &  0.094 & 2.08   \\
LLaVA-Chef  & $X_i$ + $X_{ing}$ & 0.337 & 0.197 & 0.12 & 0.077 & 0.156 & 0.177 & 0.45 & 0.156 & 0.232 & 0.203 & 2.43   \\
 
LLaVA-Chef  &  $X_i$ + $X_t$ + $X_{ing}$ & 0.366 &  0.218  & 0.137 & 0.09 &  0.170 &  0.189 &   \textbf{0.473} &  0.17 &  0.240 &   0.242 &  17.90  \\

\hline

Phi-3  &$X_t$ &  0.178  & 0.089  & 0.047  & 0.025  & 0.029  & 0.187   & 0.268  & 0.069  & 0.134  & 0.003 & 11.51 \\
Phi-3  &  $X_{ing}$&  0.202  & 0.108  & 0.06  & 0.034  & 0.039  & 0.207   & 0.298  & 0.087  & 0.149  & 0.003 & 11.65 \\
Phi-3  &$X_t$ + $X_{ing}$&  0.209  & 0.114  & 0.064  & 0.038  & {\bf 0.216}  & 0.042   & 0.31  & 0.095  & 0.155  & 0. & 11.99 \\

\hdashline

 \textit{KERL-Recipe} &$X_t$  & 0.292  & 0.155  & 0.089  & 0.053 & 0.072 & 0.132  & 0.317  & 0.09  & 0.171  & 0.10 &   7.60 \\
\textit{KERL-Recipe}  & $X_{ing}$& 0.392  & 0.249  & 0.170  & 0.12 & 0.15 & 0.188  & 0.441  & 0.179  & 0.239  & 0.323 &  8.29  \\
\textit{KERL-Recipe}  &$X_t$ + $X_{ing}$& \textbf{0.405}  & \textbf{0.257}  & \textbf{0.175}  & \textbf{0.123} & 0.154 & \textbf{0.195}  & 0.454  & \textbf{0.183}  & \textbf{0.241}  & \textbf{0.347} &  7.68  \\
\hline

\end{tabular}
\end{adjustbox}
\caption{\small Performance on Recipe Generation} 
\label{tab:recipe_gen}
\end{table*}

\begin{table*}[!ht]
    \centering
 \begin{adjustbox}{width=0.98\textwidth}
    \begin{tabular}{|l|l|c|c|c|c|c|c|c|c|c|c|}
\hline

Model & Inputs & Calories & Fat Calories & Protein &  Sugar & Fiber & Carbohydrates & Sodium & Cholesterol & Saturated Fat  & Total Fat \\
\hline
Dataset Mean &  & 426.39$\pm$ 626.7 & 189.79$\pm$332.39 & 15.83$\pm$21.26 & 18.56$\pm$ 52.27 & 3.55$\pm$ 5.56 & 43.64$\pm$ 80.57 & 0.66$\pm$ 2.03 & 0.08$\pm$ 0.14& 8.15$\pm$ 16.49& 21.14$\pm$36.93 \\
\hdashline

LLaVA-Chef  & $X_t$ & 306.54 & 172.16 & 11.8 & 17.44 & 3.85 & 33.31 & 29.02 & 3.13 & 10.46 & 23.4 \\

LLaVA-Chef  &  $X_{ing}$ &  323.75 &  160.45 & 15.47 & 20.96  & 7.48  & 38.39  & 160.72  & 16.82   &  19.81 & 37.61 \\
LLaVA-Chef  & $X_{instruct}$ & 319.42  &  161.77  & 15.39 &  22.31 & 7.86  & 38.87  & 111.78  & 14.63   & 19.76  & 36.94   \\
LLaVA-Chef  & $X_{t}$ + $X_{ing}$ + $X_{inst}$ & 323.6 & 161.5 & 12.68  & 20.6 & 5.43 & 39.29  & 192.55 & 21.51 & 19.78 & 40.32 \\

\hdashline

\textit{KERL-Nutri}   & $X_t$ & 258.28 &  134.81 & 8.97 & 14.22 & 2.32 & 29.05 & 0.5 & 0.05  & 6.12 & 14.98  \\

\textit{KERL-Nutri}   & $X_{ing}$ & 226.65 &  104.98 & 7.44 & 12.28 & 1.88 & 25.38 & 0.38 & 0.04  & 4.56 & 11.67  \\

\textit{KERL-Nutri}   & $X_{instruct}$ & 245.54  &  124.7  & 8.58 & 13.58  & 2.19  & 27.68  & 0.46  &  0.04  & 5.45  &  13.86  \\

\textit{KERL-Nutri}   &$X_{t}$ + $X_{ing}$ + $X_{inst}$  & \textbf{221.38} & \textbf{103.03} & \textbf{7.29} & \textbf{11.92} & \textbf{1.84} & \textbf{24.69} & \textbf{0.37} & \textbf{0.04} & \textbf{4.48} & \textbf{11.45} \\

\hline

\end{tabular}
\end{adjustbox}
\caption{\small Performance on Nutrient Generation: Mean absolute error per micro-nutrient. }
\label{tab:nutri19K_gen100}
\end{table*}



\subsection{\textit{KERL-Recipe} Evaluation}
Given recipe titles $X_t$, recipe ingredients $X_{ing}$ and recipe images $X_i$, or subsets thereof, we now evaluate how well models can generate the actual recipe cooking steps $X_{inst}$.
We compare pretrained LLMs and their fine-tuned counterparts for recipe generation in Table~\ref{tab:recipe_gen}. LLaVA-Chef \cite{LlaVAChef}, based on LLaVA, is a state-of-the-art model for this task, and it shows better scores than the pre-trained LLaVA and LLaMA baselines. However, we can observe that \textit{KERL-Recipe}, not only outperforms its base Phi-3 model, it has the best overall performance along various recipe quality metrics. It outperforms LLaVA-Chef, which involves entire model training, whereas \textit{KERL-Recipe} is only LoRA fine-tuned. LLaVA-Chef improves almost 7 points on BLEU-1 over its base LLaVA model, whereas \textit{KERL-Recipe} improves about 20 points over Phi-3. Both LLaVA-Chef and \textit{KERL-Recipe} perform better when provided with ingredients $X_{ing}$, compared to only using the recipe title ($X_t$), suggesting that the ingredients are important in recipe generation. Note that LLaVA-Chef has the ability to process images, which enables it to generate recipes from food images, whereas \textit{KERL-Recipe} is a text-only model. Overall, \textit{KERL-Recipe} not only outperforms in most metrics, it requires training fewer parameters (LoRA adapter), yet improves over its base model with significant margin.

\subsection{\textit{KERL-Nutri} Evaluation}

Nutrition generation module (\textit{KERL-Nutri}) is based on LoRA fine-tuning the base Phi-3 model. For comparison, we also fine-tune LLaVA-chef (the full model) on the nutrition generation benchmark. Our model outperforms others, as evident in Table \ref{tab:nutri19K_gen100} where the first row shows the mean values of the micro-nutrients in the test set. LLaVA-Chef estimates nutrition slightly better when only title $X_t$ is given compared to using ingredients $X_{ing}$ or instructions $X_{instruct}$.  However for \textit{KERL-Nutri}, ingredients play a crucial role in nutrition estimation, as they contain the actual nutrients.  Generating nutrients from only instruction has slightly higher MAE, as instructions may not explicitly mention all ingredients. Overall, \textit{KERL-Nutri} achieves lower errors when provided with the complete recipe, including the title, ingredients, and cooking instructions.

\section{Conclusion}

We present KERL, a food recommendation system that combines the power of KGs with LLMs in a question answering framework. We also create a large-scale QA benchmark dataset using FoodKG.  After evaluation of several open source LLMs we selected Phi-3-mini as the base LLM, training it to understand the subgraphs from FoodKG to help answer complex constrained questions regarding personalized food recommendations. 
Using a multi-LoRA approach, we also fine-tune adapters to generate cooking steps and nutritional information for the recipes, offering a seamless solution for meal planning and cooking.  Our evaluation shows that KERL outperforms baseline models for all three tasks, with more relevant recipes, better quality cooking steps, and more accurate nutrient values. In the future, we plan to leverage Chain-of-Thought reasoning along with RAG to further improve the performance while incorporating ingredient substitution, person's health information, and cultural preferences. 


\section{Limitations}
\begin{itemize}
    \item \textit{KERL-Recom} relies on the recipe subgraphs retrieved from FoodKG \cite{haussmann2019foodkg}. Therefore, the system will not recommend any recipe if none of the recipe in KG meet all the constraints. The system may also fail if incorrect context information is provided, hence the results should not be used without proper safeguards.

    \item  \textit{KERL-Recom} do not directly establish the relationship between the person's health conditions and the corresponding dietary restrictions. For example, it can recommend sugar-free recipes, but it may not accurately recommend the correct recipes for a diabetic person. This capability is left for future research.
   
    \item \textit{KERL-Nutri} generates the micro-nutritional information for most recipes, but it may not accurately generate micro-nutritional details for extreme cases, such as for recipes with high or very low calories. 
    
\end{itemize}

\bibliography{acl}

\clearpage
\newpage

\appendix
\label{sec:appendix}


\section{Dataset Details}
\label{apendx:dataset_details}

\subsection{Recipe tags} 
\label{appendx-tags}
\begin{figure*}[ht]
    \centering
    \includegraphics[scale=0.39]{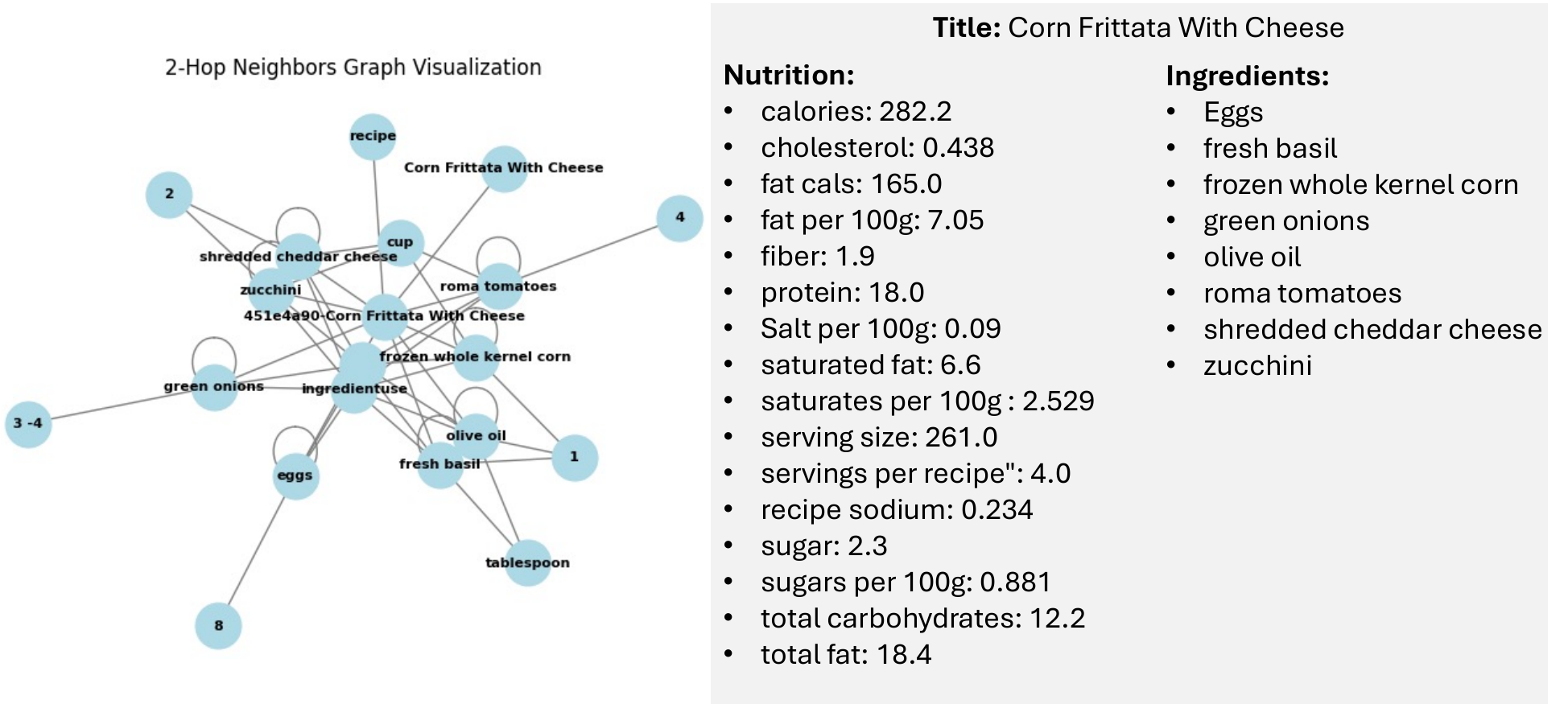}
    \caption{\small FoodKG Recipe sample: left panel shows a 2-hop KG subgraph for the recipe node shown on the right.}
    \label{fig:sample1}
\end{figure*}

Recipes in FoodKG are tagged with one or more tags, making a total of 490 unique tags.
We selected recipes associated with health-related tags, listed in Table~\ref{tab:tags}, for the generation of KGQA benchmark to ensure that the question-answer pairs inherently focus on health constraints. Note that the recipes tagged with these tags were also tagged with 454 other tags, indicating that the dataset covers a wide variety of recipe types. An example of a KG subraph correponding to a recipe is shown in Fig.~\ref{fig:sample1}.

\begin{table}[hbt]
    \centering
    \begin{adjustbox}{width=0.45\textwidth}
    \small
    \begin{tabular}{|l l l|}
    \hline
    lactose & vegan & vegetarian \\
    dairy-free & gluten-free &  nut-free \\
    egg-free & low-carb & low-fat \\
     low-sodium & low-cholesterol & low-protein  \\
      high-protein & high-calcium &  high-fiber  \\
        \hline
    \end{tabular}
    \end{adjustbox}
    \caption{\small Tags representing various dietary preferences and nutritional constraints.}
    \label{tab:tags}
\end{table}


\subsection{KGQA benchmark details} 
\label{apendx:sub_kgqa_details}
The recipes in FoodKG, the knowledge base used for KGQA benchmark, are tagged with one or more tags. 
Table \ref{tab:kgqa_details} shows the number of tagged recipes for each tag used for dataset generation and the associated number of questions in the train and test splits. The limited number of dairy-free tagged recipes (only 7) led to fewer corresponding test questions (only 3). As a result, the evaluated models also show the lowest F1 score for this tag, as shown in Figure \ref{fig:per_tag_f1}.

\begin{table}[hbt]
    \centering
    \begin{adjustbox}{width=0.45\textwidth}
    \small
    \begin{tabular}{|l|c|c|c|}
    \hline
    Tag & Tagged recipes & Train set &   Test set\\
      \hline
      lactose &366& 874 &112 \\
      vegan&968& 2287 &314\\
      vegetarian &3392&8142& 979\\
      dairy-free &7&18 &3\\
      gluten-free &565&1328&187 \\
      nut-free &45& 107 &13 \\
      egg-free &440&1078 &124 \\
      low-carb &4239& 10248 &1244\\
      low-fat&2202& 5296&639\\
      low-sodium&4445& 10561&1407\\
      low-cholesterol&3710&8990& 1059\\
      low-protein&3320&7944&1032 \\
      high-protein&690&1642&219 \\
      high-calcium&540&1318&162 \\
      high-fiber&33&76 &8\\
      \hline
    \end{tabular}
\end{adjustbox}
    \caption{\small KGQA benchmark: Number of tagged recipes for each tag and questions for each tag.}
    \label{tab:kgqa_details}
\end{table}

\section{Foundational Models}
\label{apdx-subsec:llms}
Here we provide a list of baseline LLMs we compare with in our empirical evaluation.

\textbf{internLM2} \cite{cai2024internlm2} is the second generation internLM model, trained to capture long-term dependencies. It outperforms on 30 benchmarks in long context modeling and open-ended subjective evaluations.

\textbf{Mistral} ~\cite{jiang2023mistral} is engineered for superior performance and efficiency. Its 7B model can outperforms LLaMA-2 13B model.

\textbf{LLama-2}~\cite{touvron2023llama2} is a collection of foundation language models ranging from 7B to 70B. Due to the popularity of the llama series, we select Llama-2-7B model in our study.

\textbf{Llama-3.1} ~\cite{meta2024introducing} is a set of large scale very powerful open source LLM that improves upon Llama-2, and is comparable to the flagship models like GPT-4 and Claude 3.5 Sonnet. Therefore, it became an obvious choice for our study.

\textbf{Phi-2}~\cite{mojan2023phi2} is a 2.7B parameter LLMs designed for efficient and high-performing natural language processing tasks. It has demonstrated better performance than the LLaMA-2 (13B) and Mistral (7B) models on a range of benchmark tasks, showcasing its effectiveness in various NLP domains.

\textbf{Phi-3}~\cite{abdin2024phi} has improved models in the Phi series; even its mini version with 3.8B parameters outperforms several 7B and 13B models. We used Phi-3-mini-4k and Phi-3-mini-128K in our study for their performance despite their smaller size.

\section{Metrics}
\label{apdx-subsec:metrics}
\subsection{Metrics for Recommendation Evaluation}
\label{apdx-subsec:metrics_recom}
We use standard retrieval metrics and provide their formal definitions considering order agnostic evaluation of all the models. Let $Y$ be a list of recipes as ground truth answer and $\Tilde{Y}$ a list of recipes recommended by the model. Then, we define true positive (TP), false positive (FP) and false negative (FN) as follows:

\begin{equation*}
\begin{aligned}
    TP &= Y \cap \Tilde{Y} \\
    FP &= \Tilde{Y} - Y \\
    FN &= Y - \Tilde{Y}
\end{aligned}
\end{equation*}

Then precision (P), recall (R), and F1 scores are computed as follows:

\begin{equation*}
\begin{aligned}
    P &= \frac{|TP|}{|TP| + |FP|} \\
    R &= \frac{|TP|}{|TP| + |FN|} \\
    F1 &= \frac{2PR}{P+R} \\
\end{aligned}
\end{equation*}

We compute precision at rank $r$ for all relevant recipes $M$ and average them to get average precision (AP). Then, we calculate mean average precision (mAP) by taking the average of AP across all $N$ samples, formally defined as follows:

\begin{equation*}
\begin{aligned}
    AP &=  \frac{1}{|M|} \sum_{r \in M} P(r) \\
    mAP &=  \frac{1}{N} \sum_{i=1}^N AP_i \\
\end{aligned}
\end{equation*}

\subsection{Metrics for Recipe Generation}
\label{apdx-subsec:metrics_recipe_gen}
Here we provide formal definitions of the different metrics used in our evaluation of generated recipes.

\paragraph{\textbf{BLEU score:}}
Bilingual Evaluation Understudy (BLEU) score ~\cite{papineni2002bleu}, initially proposed for machine translation evaluation, is a metric that quantifies the similarity between a generated text sequence and a reference text sequence. Let $Y_{pred}$ be the predicted text sequence of length $n_p$, $Y_{label}$ be the ground truth text sequence of length $n_l$, and N the number of $n$-grams, then the BLEU score is defined as:

\begin{equation*}
    BLEU =BP \exp \sum_{n=1}^N w_n \cdot \log (prec) \\    
\end{equation*}

\begin{equation*}
    prec = \frac{\sum\limits_{p \in Y_{pred}} \sum\limits_{n\text{-}gram \in p} Count_{clip} (n\text{-}gram)}{\sum\limits_{p^{'} \in Y_{pred}} \sum\limits_{n\text{-}gram^{'} \in p^{'}} Count (n\text{-}gram^{'})}
\end{equation*}

\begin{equation*}
BP =
    \begin{cases}
         1, & n_p > n_l \\
        e^{1- n_p/n_l} & n_p \leq n_l
    \end{cases}
\end{equation*}

Where, $prec$ is $n$-gram precision, $w_n$ is the weight for each precision score and BP is brevity penalty that penalizes too short sequences. For the BLEU-N score, the weight of the precision score is $w_n = \frac{1}{N}$. 
BLEU cannot be directly compared between research papers~\cite{post2018call} as it is a parameterized metric and the parameters are often not reported. Therefore, we adhere to the standard implementation of BLEU-N (\url{https://github.com/salaniz/pycocoevalcap}). Furthermore, SacreBLEU~\cite{post2018call} (\url{https://github.com/mjpost/sacreBLEU}) is a reproducible and shareable implementation of the BLEU score.

\paragraph{\textbf{Rouge score:}}
Rouge (Recall-Oriented Understudy for Gisting Evaluation) ~\cite{lin2004rouge} was designed to evaluate the text summarization system. Rouge-N computes N-gram recall between predicted text (summary) and ground truth, and is defined as:

\begin{equation*}
\begin{aligned}[b]
   & Rouge\text{-}N = \\ & \frac{\sum\limits_{s \in Y_{label}} \sum\limits_{n\text{-}gram \in s} Count_{match}(n\text{-}gram)}{\sum\limits_{s \in Y_{label}} \sum\limits_{n\text{-}gram \in s} Count(n\text{-}gram)}
\end{aligned}
\end{equation*}

Where $Count_{match}$ is maximum number of $n$-grams co-occurring in predicted text $Y_{pred}$ and reference ground truth $Y_{label}$. Rouge-L~\cite{lin-och-2004-automatic} operates on the basis of the longest common subsequence  between generated text and ground truth references.
It measures the extent to which the generated text captures the longest in-sequence co-occurrences of words in the references.

\paragraph{\textbf{CIDEr:}}
Consensus-based Image Description Evaluation (CIDEr) \cite{vedantam2015cider}  was introduced as a metric to quantify the quality of generated image captions or descriptions. It operates by measuring the degree of consensus between a generated caption and a set of human-authored reference captions. The mathematical formulation of CIDEr is as follows:


\begin{equation*}
\begin{aligned}[b]
   & CIDEr (Y_{pred}, Y_{label}) = \\ & \sum_{n=1}^N w_n  \frac{1}{m} \sum\limits_{j}  \frac{g^n(Y_{pred}). g^n(Y_{label}^j)}{||g^n(Y_{pred})|| ||g^n(Y_{label}^j)||}
\end{aligned}
\end{equation*}

where $Y_{label}^j$ is $j^{th}$ ground truth sentence, $g^n(.)$ is a vector representing Term Frequency Inverse Document Frequency (TF-IDF) weighting for each $n$-gram. 

\paragraph{METEOR:}
METEOR (Metric for Evaluation of Translation with Explicit Ordering) ~\cite{elliott2013image} is a machine translation evaluation metric that leverages unigram matching between machine-generated translations (hypotheses) and human-produced references (ground truth). It incorporates both precision (\textit{P}) and recall (\textit{R}) of unigrams, as well as other features such as word order and synonym matching, to arrive at a comprehensive assessment of translation quality. The formal definition of METEOR is as follows:


\begin{equation*}
    METEOR = \frac{10PR}{R+9P} (1- Penalty)
\end{equation*} 
The penalty factor accounts for word order and length differences between the hypothesis and reference.

\begin{table*}[!ht]
    \centering
 \begin{adjustbox}{width=0.98\textwidth}
 \small
    \begin{tabular}{|l|l|c|c|c|c|c|c|c|c|c|c|}
\hline

Model & Inputs & Calories & Fat Calories & Protein &  Sugar & Fiber & Carbohydrates & Sodium & Cholesterol & Saturated Fat  & Total Fat \\
\hline
Dataset Mean &  & 321.76 $\pm$ 222.82& 135.36$\pm$122.82 & 12.69$\pm$ 12.51 & 10.01$\pm$ 11.89& 2.67$\pm$ 2.54& 29.86$\pm$23.94 & 0.44$\pm$0.44 & 0.05$\pm$0.06 & 5.62$\pm$5.74 & 15.09 $\pm$ 13.65 \\
\hdashline

LLaVA-Chef  & $X_t$ & 205.13  & 118.21  & 9.22 & 9.6 & 3.07 & 20.9 & 28.66 & 2.69 & 8.07 & 17.97   \\
LLaVA-Chef  & $X_{ing}$ & 229.17 & 111.5 &  13.23 &  14.61 &  6.82 &  27.45 & 155.73 &  15.36 &  18.13 & 34.44  \\
LLaVA-Chef  & $X_{inst}$ & 222.32 & 111.05 & 13.25  & 15.34  & 7.23  & 27.38  &  110.52 & 13.5  & 17.94  & 32.85  \\
LLaVA-Chef  & $X_{t}$ + $X_{ing}$ + $X_{inst}$ & 233.35 &  113.36& 10.5  & 14.01  & 4.81  & 28.4  & 188.49 & 19.37  & 17.79  & 36.52   \\
\hdashline
\textit{KERL-Nutri}  & $X_t$ & 159.44  & 85.58  & 6.64 & 6.62 & 1.61 & 16.17 & 0.29 & 0.03 & 3.91 & 9.51 \\
\textit{KERL-Nutri}  & $X_{ing}$ & 132.62 & 61.23 & 5.47 & 4.82 & 1.22 & 12.62 & 0.21 & \textbf{0.02}  & 2.56 & 6.8 \\
\textit{KERL-Nutri}  & $X_{inst}$ & 147.19 & 75.84 & 6.29 & 6.11 & 1.46 & 14.91 & 0.26 & 0.03 & 3.27 & 8.43 \\
\textit{KERL-Nutri} &$X_{t}$ + $X_{ing}$ + $X_{inst}$  & \textbf{127.67} & \textbf{59.49} & \textbf{5.3} & \textbf{4.64} & \textbf{1.18} & \textbf{12.09} & \textbf{0.2} &\textbf{0.02} & \textbf{2.48} & \textbf{6.61} \\

\hline

\end{tabular}
\end{adjustbox}
\caption{\small Performance of nutrition generation models, filtered to include only the 95th percentile of samples. }
\label{tab:nutri_gen95}
\end{table*}

\paragraph{\textbf{Perplexity:}}
Perplexity, a widely used metric for evaluating autoregressive or causal language models, quantifies the degree of uncertainty a model exhibits when predicting the next token in a sequence. It is formally defined as the exponential average negative log-likelihood of a given text sequence. Mathematically, for a text sequence $X$ of length $m$ generated using a model  $f_{\theta} (.)$, perplexity can be calculated as:

\begin{equation*} 
    ppl (X) = exp \left({ -\frac{1}{m} \sum_{i}^{m} \log f_{\theta}}  (x_i | x_{<i})\right)
\end{equation*}
where, $f_{\theta}  (x_i | x_{<i})$ signifies the probability assigned by the model to the token $x_i$, conditioned on the preceding tokens $x_{<i}$.

\subsection{Metrics for Nutrition Generation}
\label{apdx-subsec:metrics_nutri_gen}
Micro-nutrients were formatted in a pre-defined JSON style during the training, so the model was also expected to generate text in a similar style, making it easy to parse micro-nutrients and their values. Note that each generated sample output may not contain all the desired micro-nutrients. Therefore, for each sample, we consider the micro-nutrient tags found in the generated text. We parse all the micro-nutrient tags from the generated text along with their numerical values and compute the mean average error with the ground truth, formally defined as:

\begin{equation*}
    MAE (nutri) = \frac{1}{n} \sum_{i=1}^n |y_{nutri}^i -  \Tilde{y}_{nutri}^i|
\end{equation*}
Where, $y_{nutri}^i$ and $\Tilde{y}_{nutri}^i$ are the ground truth  and predicted values of the nutrient, respectively.

\section{Additional Results}

\subsection{\textit{KERL-Recom}}



    

\paragraph{Performance on Recipe Types}
\label{apndx:results_recipe_types}

Table \ref{tab:test_8k_2} shows the performance of open source LLMs and our model on the KGQA benchmark for recipes tagged with some of the tags such as lactose, vegan, vegetarian, gluten-free and nut-free. LLaMA-2 ranks as the second best, except for the gluten-free tag. Mistral performs similarly to or slightly better than Phi-3-mini, but Phi-3-mini is smaller than the other models. Overall, \textit{KERL-Recom}, leveraging Phi-3-mini as the base model, achieves precision and F1 greater than $90$ for all tags.

\begin{table}[!htb]
\centering
 \begin{adjustbox}{width=0.495\textwidth}
    \small
    \begin{tabular}{|l|c|c|c|c|c|c|c|}
    \hline
      Model   &  Tag   & mAP & P & R & F1   \\
      \hline
      internLM2 & \multirow{7}{*}{lactose} &   0.015 & 0.008 & 0.017 & 0.011    \\
      Mistral &   & 0.14 & 0.538 & 0.368 & 0.437     \\
      Llama-2 &   &0.477 & 0.838 & 0.514 & 0.637    \\
      Llama-3.1&     & 0.152 & 0.233 & 0.321 & 0.27   \\
      Phi-3-mini-128K&   &   0.202 & 0.812 & 0.217 & 0.343    \\
      \textit{KERL-Nutri} &   & {\bf 0.898} & {\bf 0.955} & {\bf 0.916} & {\bf 0.935}  \\
      \hline
       internLM2 & \multirow{7}{*}{vegan} &    0.09 & 0.038 & 0.0079 & 0.051    \\
      Mistral &  &   0.201 & 0.549 & 0.492 & 0.519     \\
      Llama-2 & &   0.669 & 0.885 & 0.71 & 0.788    \\
      Llama-3.1&   &    0.161 & 0.296 & 0.43 & 0.351   \\
      Phi-3-mini-128K&   &    0.404 & 0.873 & 0.421 & 0.568    \\
      \textit{KERL-Nutri} &  &   {\bf 0.964} & {\bf 0.988} & {\bf 0.975} & {\bf 0.981}   \\
      \hline
      
    internLM2 & \multirow{7}{*}{vegetarian}    & 0.085 & 0.034 & 0.078 & 0.048    \\
      Mistral &   & 0.201 & 0.531 & 0.52 & 0.526    \\
      Llama-2 & &  0.639 & 0.856 & 0.708 & 0.775     \\
      Llama-3.1&     & 0.179 & 0.325 & 0.462 & 0.381 \\
      Phi-3-mini-128K&      & 0.361 & 0.871 & 0.352 & 0.501   \\
      \textit{KERL-Nutri} &  & {\bf 0.966} & {\bf 0.987} & {\bf 0.976} & {\bf 0.981}  \\
      \hline

     internLM2 & \multirow{7}{*}{gluten-free}   & 0.062 & 0.027 & 0.058 & 0.037    \\
      Mistral* &   & 0.26 & 0.581 & 0.673 & 0.624    \\
      Llama-2 &   &0.559 & 0.87 & 0.609 & 0.717    \\
      Llama-3.1&     & 0.208 & 0.364 & 0.548 & 0.438    \\
      Phi-3-mini-128K&     & 0.282 & 0.811 & 0.285 & 0.421  \\
      \textit{KERL-Nutri} &    & {\bf 0.939} & {\bf 0.982} & {\bf 0.951} & {\bf 0.966}  \\
      \hline

    internLM2 & \multirow{7}{*}{nut-free}   & 0.103 & 0.019 & 0.067 & 0.029     \\
      Mistral* &    & 0.224 & 0.542 & 0.591 & 0.565    \\
      Llama-2 &   & 0.628 & 0.833 & 0.682 & 0.75     \\
      Llama-3.1&     & 0.243 & 0.345 & 0.455 & 0.392   \\
      Phi-3-mini-128K&     & 0.385 & 0.786 & 0.367 & 0.5   \\
      \textit{KERL-Nutri} &   & {\bf 1.0} & {\bf 0.909} & {\bf 1.0} & {\bf 0.952}   \\
      \hline
      
    \end{tabular}
    \end{adjustbox}

    \caption{\small Results on KGQA test set reported for several tags. Overall, \textit{KERL-Recom} performs better for numerous types of recipes.}
    \label{tab:test_8k_2}
\end{table}

\textbf{Qualitative results}
Despite the impressive performance of \textit{KERL-Recom}, it is prone to failure by recommending false positives or missing true positive in recommended recipes. For examples, row-2 in Table \ref{tab:recom_samples} shows the question where \textit{KERL-Recom} suggested false positive whereas in row-3 it suggested a recipe that is not even in context. Similarly, the last row demonstrates an example, where the model failed to select all true positives from context, resulting few true negatives.

\begin{table*}[ht]
\begin{adjustbox}{width=0.97\textwidth}
    \centering
    \begin{tabular}{|p{6cm}|p{4.5cm}|p{4cm}|}
    \hline
        User question & Recipe in context &  \textit{KERL-Recom} recommendations \\
        \hline
       What low-protein recipes use crushed red pepper flakes, bacon, ginger ale, ground black pepper, pepper and exclude cream of coconut, tamari, fresh thyme leaves, and have fiber more than 4.28, sugars per 100g within range (0, 5.99)? &
       \multirow{5}{4cm}{Tamarind Juice, \\ \color{blue}Mock Sangria, \\ Low-Carb Balsamic Dressing,\color{black} \\ County Cherry Dessert, \\ \color{blue}Mock Champagne \color{black}}
            &  \multirow{3}{4cm}{Low Carb Balsamic Dressing, \\ Mock Champagne, \\ Mock Sangria} \\
                        \hdashline
                        
         Suggest me vegetarian dishes that require fresh ground black pepper, green onions, fresh parsley, plain yogurt, red onions and must not contain fine salt, peach slices, arhar dal, and have a total of fiber not above 5.4, sugars per 100g no more than 3.66.  &   
           \multirow{3}{4cm}{Gyoza or Pot Sticker Dipping Sauce,\\
           \color{blue} Wild Rice Stuffing Side Dish, \\ \color{black}
            Pixie Cookies} & 
             \multirow{2}{4cm}{Wild Rice Stuffing Side Dish, \\
            Pixie Cookies}\\

            \hdashline
       Can you list the low-carb recipes that use curry powder, cooked spaghetti, steak, bottled hot pepper sauce, condensed beef broth but do not contain whole wheat pancake mix, chicken thigh fillets, white bread, while containing fat cals not less than 203.0, and protein within range (0, 32.5)? &  
       
        \multirow{3}{4cm}{Quick Sausage, White Bean and Spinach Stew, \\
           \color{blue} Jamaican Brown Stew Chicken,\color{black} \\
            Watermelon, Cucumber and Feta Salad} &  
            
            \multirow{2}{4cm}{ Jamaican Brown Stew Chicken, \\
            Manic Bullet} \\

    \hdashline
    Can you suggest low-sodium recipes cooked with fresh ground black pepper, plain yogurt, all - purpose flour, fresh lemon juice, apples but do not have garam masala, wheat and have protein no more than 13.08, sugars per 100g no more than 11.53? & 
    
    \multirow{7}{4cm}{Fudge Pie, Pasta Pascal,\\
            \color{blue} Chocolate-Pecan Brownies,\\
            Cold Oven Pound Cake,\color{black}\\
            Cucumber and Feta Salad, \\
           \color{blue} Lemon Meringue Tart,\color{black} \\
            Cold Oven Pound Cake} &  
             \multirow{3}{4cm}{Fudge Pie, \\
            Cold Oven Pound Cake, \\
            Lemon Meringue Tart}\\
         \hline
    \end{tabular}
\end{adjustbox}
    \caption{
    Qualitative results of KERL-Recom: The second column lists the recipes in the subgraph (only names for simplicity) where blue color shows recipes that satisfy the user constraints $R^+(t_j)$. Row 1 shows a perfect result, row 2 shows one false positive recommendation, row 3 shows two suggested recipes, with only one present in the context and is also true positive, and the final row shows a subset of the ground truth selected by the model with missing true positives from recommended recipes. 
    These sample results suggest that despite showing strong performance, it may fail by suggesting false negatives, missing true positives, or recommending unrelated recipes.
     }
    \label{tab:recom_samples}
\end{table*}

\subsection{\textit{KERL-Nutri}}
\label{apdx:nutri_estimation}


In Table~\ref{tab:nutri19K_gen100}, we compare the performance of LLaVA-Chef and our \textit{KERL-Nutri} model on generating the  micro-nutrients for the recommended recipes. For some nutrients, the MAE is rather large. This happens because some ground truth samples have abnormally high or zero nutritional values, introducing noise that affects model performance. To analyze this, we filtered the samples within a specific percentile range, excluding outliers, and then calculated the MAE. Detailed results on nutrient generation for the 95th percentile of samples are shown in Table \ref{tab:nutri_gen95}.  We observe that removing noisy samples reduces MAE by about half for all the nutrients. Nevertheless, our \textit{KERL-Nutri} remains the superior model.

\end{document}